\pdfoutput=1
\documentclass[11pt]{article}

\usepackage[final]{acl}

\usepackage{hyperref}
\usepackage{url}

\usepackage[pdftex]{graphicx}
\usepackage{color, colortbl}
\definecolor{Gray}{gray}{0.9}
\usepackage{booktabs}
\usepackage{times}
\usepackage{soul, color, xcolor}
\usepackage[utf8]{inputenc}
\usepackage{algorithm}
\usepackage{multirow} 
\usepackage{algpseudocode}
\usepackage{booktabs}
\usepackage{latexsym}
\usepackage{amsmath}
\usepackage{color}
\usepackage{xcolor} % 引入xcolor包
\usepackage{soul} % 引入soul包
\usepackage{array}
\usepackage{geometry}
\usepackage{multicol} 
\usepackage{xspace}
\usepackage{makecell}
\definecolor{lightyellow}{rgb}{1.0, 1.0, 0.6157}

\newcommand{\hlyellow}[1]{{\sethlcolor{lightyellow}\hl{#1}}}

\definecolor{custompurple}{rgb}{1.0, 0.7686, 1.0}
\definecolor{customgreen}{rgb}{0.65098, 0.9451, 0.65098}

\newcommand{\hlcustpurple}[1]{{\sethlcolor{custompurple}\hl{#1}}}
\newcommand{\hlgreen}[1]{{\sethlcolor{customgreen}\hl{#1}}}

\newcommand{\ourapproach}{\textsc{AMPO}\xspace}

\newcommand{\up}[1]{\textcolor{brown}{{$\uparrow$ #1}}}

\usepackage[T1]{fontenc}
\usepackage[utf8]{inputenc}

\usepackage{microtype}

\usepackage{inconsolata}

\usepackage{graphicx}

\title{AMPO: Automatic Multi-Branched Prompt Optimization}

\author{Sheng Yang\textsuperscript{1}\thanks{Equal contribution.}\thanks{This work was done during an internship at Microsoft.}\quad Yurong Wu\textsuperscript{1}\samethanks[1]\samethanks[2]\quad Yan Gao\textsuperscript{2}\thanks{Corresponding author.}\quad Zineng Zhou\textsuperscript{3}\samethanks[2]\quad Bin Benjamin Zhu\textsuperscript{2}\\ 
\textbf{Xiaodi Sun\textsuperscript{2}\quad Jian-Guang Lou\textsuperscript{2}\corresponding\quad Zhiming Ding\textsuperscript{1}\quad Anbang Hu\textsuperscript{2}\quad}\\ \textbf{Yuan Fang\textsuperscript{2}\quad Yunsong Li\textsuperscript{2}\quad Junyan Chen\textsuperscript{2}\quad Linjun Yang\textsuperscript{2}}
\\
\textsuperscript{1}Institute of Software, CAS \& University of Chinese Academy of Sciences\\
\textsuperscript{2}Microsoft\\
\textsuperscript{3}Institute of Computing Technology, CAS \& University of Chinese Academy of Sciences\\
\{yangsheng22, wuyurong20, zhouzineng22\}@mails.ucas.ac.cn; \\ \{yan.gao, binzhu, sunstifler, jlou, anbhu, juliefang, yunsongli, \\Junyan.Chen, Yang.Linjun\}@microsoft.com; zhiming@iscas.ac.cn\\
}

\begin{document}
\newcommand*\samethanks[1][\value{footnote}]{\footnotemark[#1]}
\newcommand*\corresponding[1][3]{\footnotemark[#1]}

\maketitle
\begin{abstract}
%Recognizing and managing various patterns is crucial when addressing complex issues. Typically, a human expert leverages specialized knowledge to distill patterns from data and inject relevant solutions to optimize the prompt. 
Prompt engineering is very important to enhance the performance of large language models (LLMs).
When dealing with complex issues, prompt engineers tend to distill multiple patterns from examples and inject relevant solutions to optimize the prompts, achieving satisfying results.
However, existing automatic prompt optimization techniques are only limited to producing single flow instructions, struggling with handling diverse patterns. In this paper, we present \ourapproach{}, an automatic prompt
optimization method that can iteratively develop
a multi-branched prompt using failure cases as
feedback. 
%to iteratively identify multiple patterns from failure cases and develop a more adaptable prompt structures. 
%explicitly converts a prompt into a multi-branched format and then iteratively refines the multi-branched prompt using failure cases as feedback. 
Our goal is to explore a novel way of structuring prompts with multi-branches to better handle multiple patterns in complex tasks, for which we introduce three modules: \textit{Pattern Recognition}, \textit{Branch Adjustment}, and \textit{Branch Pruning}. 
In experiments across five tasks, \ourapproach{} consistently achieves the best results. Additionally, our approach demonstrates significant optimization efficiency due to our adoption of a minimal search strategy.
\end{abstract}
\section{Introduction}
Prompt Engineering involves creating the best possible prompts to enhance the performance of large language models (LLMs). It has become increasingly significant in the development of LLM applications \cite{brown2020language, welleck2022generating}. Creating suitable prompts frequently demands considerable human effort, specialized knowledge, and numerous trial-and-error attempts \cite{zamfirescu2023johnny}. Therefore, investigating efficient automatic prompt engineering techniques is crucial \cite{Zhang_Wang_Zhou_Schuurmans_Gonzalez_2022,Chen_Chen_Goldstein_Huang_Zhou_2023}.

In recent times, automatic prompt optimization methods based on LLMs have seen extensive exploration \cite{zhou2022large,yang2023large,schnabel2024prompts}. These studies generally employ LLMs as prompt optimizers to progressively enhance prompts for target models. Particularly, one promising paradigm is feedback-based optimization methods, where LLM-based prompt optimizers act like human experts by analyzing and fixing failed cases \cite{pryzant2023automatic,wang2023promptagent,ye2023prompt}. These methods explicitly leverage the self-reflection ability of LLMs for prompt refinement and mirrors the steps of ``gradient'' in the direction of error descent and then ``propagate'' to the prompt. Currently, this paradigm of prompt optimization has achieved promising advancements and widespread interest.

\begin{figure}
    \centering
    \includegraphics[width=\columnwidth]{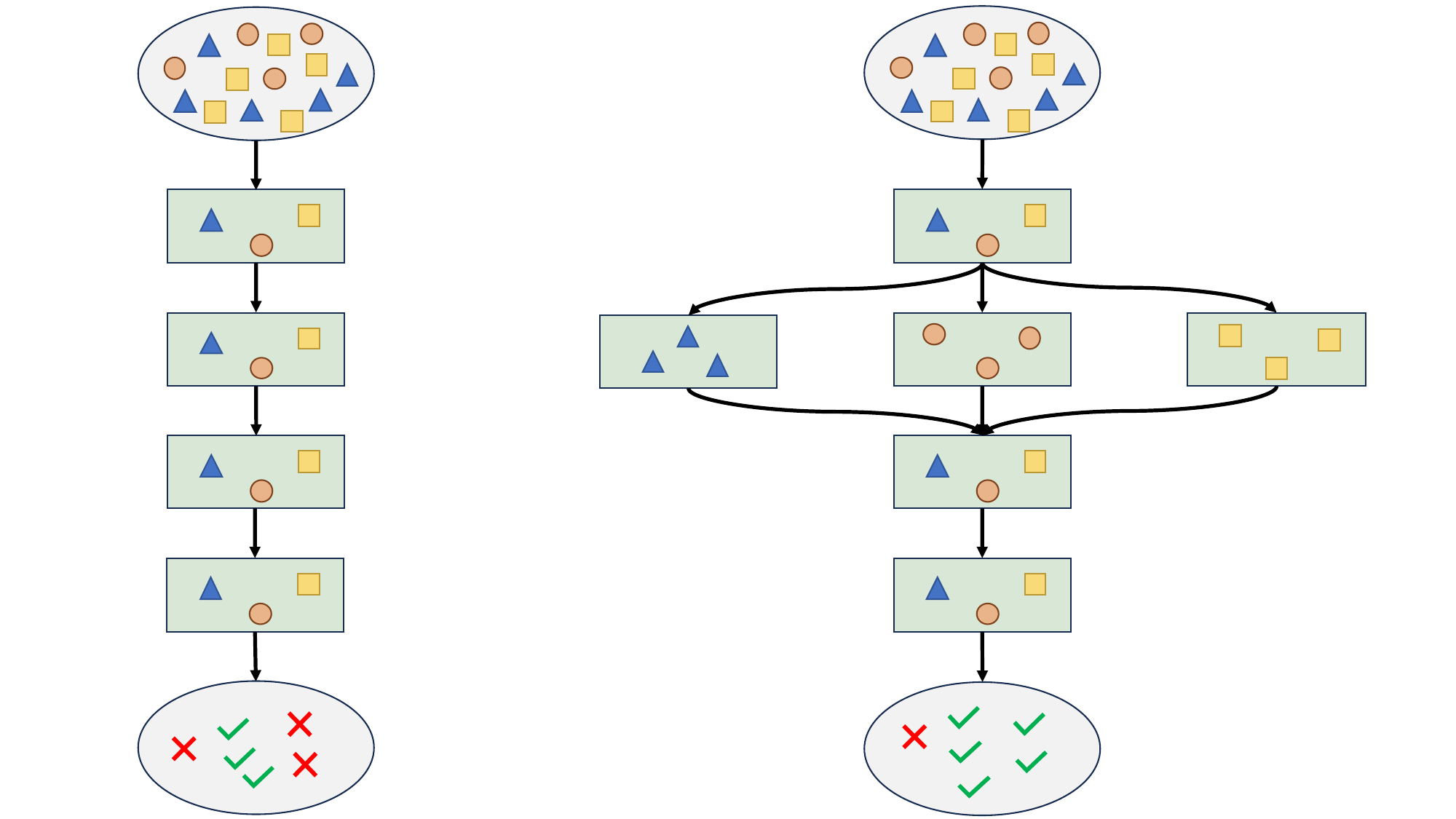}
    \caption{Our prompt optimization approach aims to iteratively optimize a single flow of instructions into a multi-branched format to handle multiple patterns.}
    \label{fig:compare}
     \vskip -0.2in
\end{figure}

Despite the success of feedback-based optimization methods, we find that the optimized prompts are primarily achieved by meticulously rewriting certain steps, attempting to provide more details. Essentially, they are mainly a form of single flow instructions. 
%because all LLM optimizers are based on optimization to existing solution by default.
 As shown in Figure \ref{fig:compare}, when dealing with complex issues, categorizing problems in advance can help LLM analyze issues more systematically and easier to find appropriate solutions
\cite{thomas2023large,ren2024representation,mao2023large}. 
It is evident that an LLM optimizer generating a single flow instructions struggles with handling various patterns. For example, current feedback-based methods either fail to provide effective fixes or cause numerous regressions when failed cases are not addressed by the single flow instructions \cite{ma2024large}. 

Given the complexity of real-world scenarios, we contend that a single flow of instructions is inadequate. The LLM optimizer should identify multiple patterns from failure cases, which in turn allows for the exploration and development of more adaptable prompt structures. Motivated by this, we propose \ourapproach{}, a prompt optimization method that moderately transforms a prompt into a multi-branched format and then optimizes the prompt using failure cases as a guide. 
The key aspect of this process is to create an adaptive prompt structure that aligns with the complexity of the task. To achieve this, we introduce three modules:
(1) \textit{Pattern Recognition} is responsible for analyzing the pattern of the failure cases. 
(2) \textit{Branch Adjustment} can adaptively choose between adding the branches to address the new pattern and providing more details to enhance the existing branches.
(3) \textit{Branch Pruning} takes both pre-pruning and post-pruning techniques to prevent the prompt from overfitting.
The advantage of our multi-branched prompt structure is that it allows the target LLM to autonomously determine the most appropriate path for each sample during inference, thereby enhancing its ability to handle diverse patterns and complexities.

We conduct experiments across five tasks, encompassing various levels of complexity.
Our experimental results show that \ourapproach{} consistently achieves the best results across five tasks from General NLU and Domain Knowledge. Additionally, compared with other feedback-based optimization approaches, our approach exhibits impressive optimization efficiency due to our adoption of a minimal search strategy. 

Our contributions are as follows:

(1) We introduce \ourapproach{}, an automatic prompt optimization method that can iteratively construct a multi-branched prompt using failure cases as feedback. Notably, \ourapproach{} is currently the first-known prompt optimization method designed to generate multi-branched prompt.

% (2) We are the first, to our knowledge, that propose using tree-structured prompt to dealing with complex issues.

% (2) To the best of our knowledge, we are the first to study the adaptivity of prompt structure.
% Our results show that \ourapproach{} can automatically find the \textit{golden mean} of multi-branched prompt from the data distribution, thereby accommodating tasks of varying
% difficulties. 

(2) To the best of our knowledge, we are the first to explore structuring prompts with multi-branches to better handle multiple patterns in complex tasks. Our results show that \ourapproach{} can automatically construct an effective multi-branched prompt from the data distribution, thereby accommodating complex tasks.

(3) We conduct experiments across five tasks. \ourapproach{} consistently outperforms other  feedback-based prompt optimization methods. Meanwhile, experimental results show that \ourapproach{} exhibits impressive optimization efficiency due to our adoption of a minimal search strategy.

\section{Motivating Example}
To illustrate the importance of recognizing and handling different patterns when dealing with complex issues, we take search query understanding task as an example, 
In this task, we employ LLM to predict a personalized query intent using the current query and user’s search history as input. As the saying goes ``there are a thousand Hamlets in a thousand people's eyes'', predicting personalized intents is quite difficult.
According to expert experience, the best approach is to first categorize users' search behaviors into distinct classes and then develop a strategy for predicting personalized intent for each category.  For example, if the current query is a \textit{re-finding query}—identical to a previous query—the most likely intent is to revisit the same webpage from the user's recent clicks.  If the current query is a \textit{reformulation query} derived from any previous queries, the intent should be personalized based on the nature of the reformulation, such as an expansion or filtering of the original query.  When the query is a \textit{domain preference query}, where the user's history indicates a preference for specific domains related to the current query, we should refine the query intent using this domain preference.  Inspired by this, \ourapproach{} introduces multiple branches to handle various categories, adaptively expanding or pruning them to achieve the most appropriate coverage.

% The Revisor can either enhance the existing steps by incorporating the Analyzer's suggestions (vertical depth) or add new branches to address new patterns (horizontal expansion). 

% Finally, we discuss principles for choosing between COT prompts and tree structure prompts, as well as issues encountered in constructing tree structure prompts, such as susceptibility to over-fitting and redundant branches.  We analyze the causes of these problems and discuss potential solutions in detail.

\section{Related Work}
\subsection{The Structure of Prompting Instructions}
Prompts significantly influences the performance of LLMs \cite{white2023prompt, liu2023pre}. Numerous studies confirm that breaking down questions into multiple intermediate steps can markedly enhance the quality of outputs \cite{wei2022chain}, particularly in tasks requiring reasoning \cite{wang2024chain,yasunaga2023large} and planning \cite{wang2023plan, zhang2023planning}. To further improve these intermediate steps, many researchers utilize the model's in-context learning abilities \cite{kojima2022large, shum2023automatic}. While these prompt design methods enhance LLMs' reasoning and generation capabilities, they still rely on a sequential, single-flow structure, which remains inadequate for more complex situations. In this work, we want to explore a novel way of structuring prompts with multi-branches to better handle multiple patterns in complex tasks. We hypothesized that a multi-branched prompt structure could allow the target LLM to autonomously determine the most appropriate path for each sample during inference.

% or employ advanced tree search techniques \cite{yao2024tree, long2023large, morris2023tree,saha2023branch}. For more complex tasks, some studies adopt recursive trees to facilitate effective interaction with external environments \cite{cao2023robot,sarthi2024raptor,saha2023branch}. Our approach extends this concept by adapting to varied and complex data distributions through the construction and optimization of multi-branched prompt.

\subsection{Automatic Prompt Optimization}
Advances in LLMs have led to numerous studies exploring automatic prompt optimization technologies \cite{ye2023prompt, schnabel2024prompts}. These studies often fall into two categories: using search algorithms \cite{guo2023connecting, zhou2022large} or leverage the self-reflection capabilities of LLMs \cite{pryzant2023automatic, yang2023large} to identify optimal prompts. Recently, there have been attempts to combine both approaches by employing search algorithms like Monte Carlo tree search (MCTS) integrated with self-reflection capabilities \cite{wang2023promptagent}. In this combined approach, each prompt is treated as a state, and each optimization as an action, enabling more refined prompt optimization through traceable tree search. Our approach falls under the self-reflection category. The key difference between \ourapproach{} and existing methods is that \ourapproach{} considers the prompt's structure as an optimization objective. Traditional methods typically treat prompts as a single block of text, optimizing them as a whole. In contrast, \ourapproach{} identifies multiple patterns from error cases and refines the prompt into a multi-branched structure by adding new branches or enhancing existing ones. This multi-branched structure enables LLMs to handle tasks with highly diverse data distributions more effectively.

% \subsection{Automatic Prompt Optimization}
% Recent advances in LLMs are prompting numerous studies to utilize these models for prompt optimization \cite{yang2023large,ye2023prompt,schnabel2024prompts}. These studies commonly employ search algorithms \cite{guo2023connecting, zhou2022large} or utilize the self-reflection capabilities \cite{pryzant2023automatic, wang2023promptagent} of LLMs to identify globally optimal prompts. Additionally, some studies aim to optimize prompts with more complex programmatic structures such as meta-prompts \cite{fernando2023promptbreeder,ye2023prompt}. However, these methods often exhibit limited effectiveness when faced with uneven or highly diverse data distributions. In response, some studies are now exploring ensemble algorithms such as AdaBoost \cite{freund1997decision} and Bagging \cite{breiman2001random} to optimize prompts by incorporating insights or examples from the ensemble models \cite{zhang2024prefer,pitis2023boosted}. In this paper, we build a multi-branched prompt to handle different patterns by iteratively adding and adjusting branches with failure cases as feedback.

\section{Methodology}

Given a task $\tau$ specified by a provided training set as $D_{{train}} = \{(q_1,a_1), (q_2,a_2)...,(q_n,a_n)\}$, where $q_i/a_i$ are input/output pairs from each entry, 
we start with an initial prompt $P_0$.
The model input consists of $P$ and $q_i$, and the base LLM $\beta$ makes the prediction based on $p_{\beta}(a_i|q_i, P)$.
The goal of prompt optimization is to find the optimal prompt $P^*$ that maximizes the performance towards a measure function $R$ (e.g., accuracy) over $D_{train}$. This can be formally defined as an optimization problem: $ P^*=argmax_{P\in S}{\textstyle \sum_{i}^{}} R(p_\beta (a_i|q_i, P))$, where $S$ denotes the sample space for a natural language prompt.
\begin{figure}
    \centering
    \includegraphics[width=\linewidth]{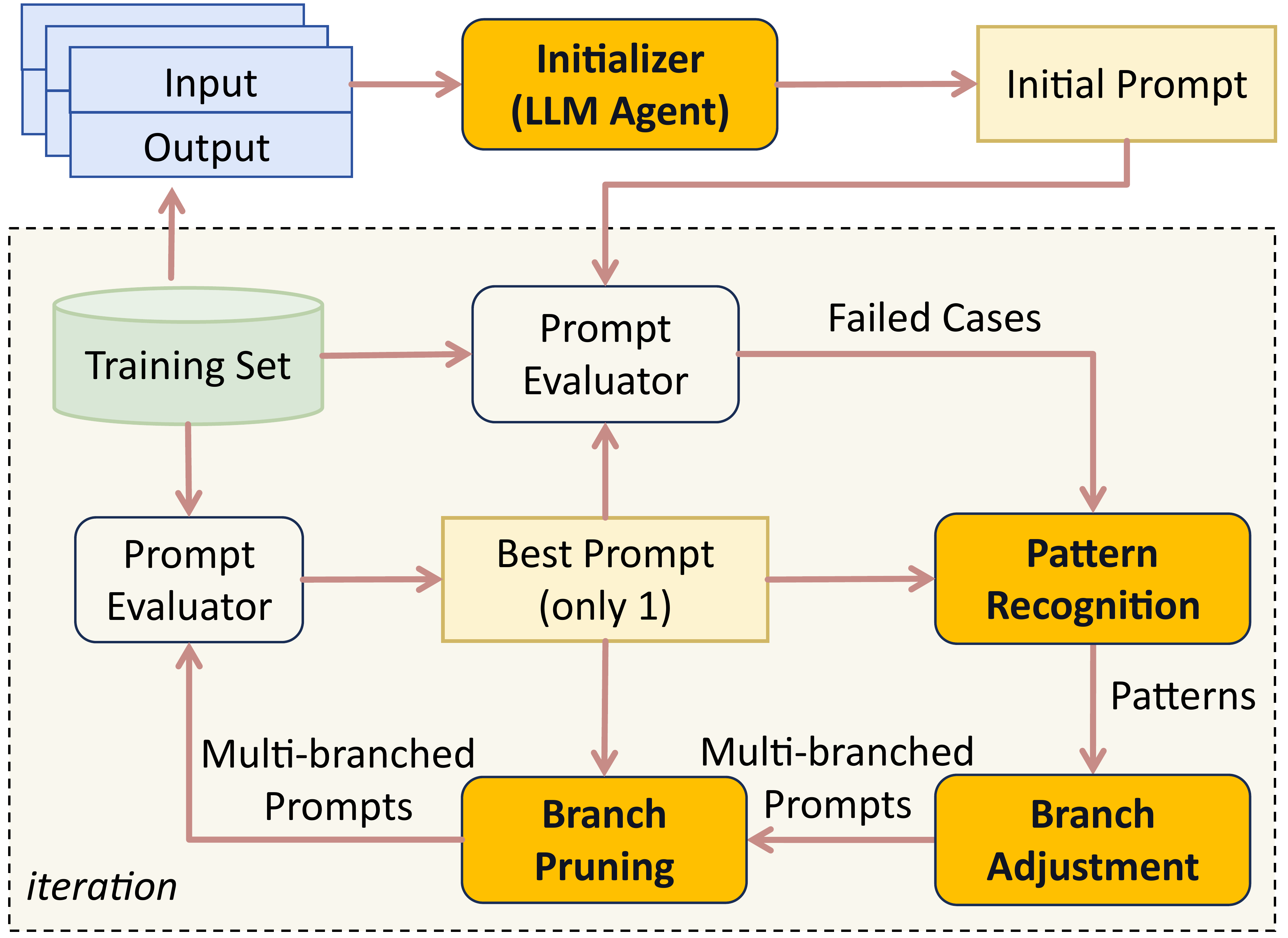}
    \caption{Overall framework of \ourapproach{}}
    \label{fig:tot}
     \vskip -0.15in
\end{figure}
\subsection{Overview}
The goal of \ourapproach{} is to iteratively develop a multi-branched prompt using failure cases as feedback. 
The key aspect of this process is generating an \textit{adaptive prompt structure} that aligns with the complexity of the task.
To achieve this, our approach mainly consists of three modules:
(1) \textit{Pattern Recognition} is responsible for analyzing the patterns of failure cases. 
% (2) \textit{Branch Adjustment} is responsible for ensuring that the \textit{golden mean} between adding the branches to address the new pattern and enhancing the existing branches according to the recognized patterns.
(2) \textit{Branch Adjustment} is responsible for balancing the adaptability between adding new branches to address emerging patterns and enhancing existing branches based on recognized patterns.
A proper balance is crucial for adaptive prompt structure. 
(3) \textit{Branch Pruning} takes both pre-pruning and post-pruning techniques to prevent the prompt from overfitting, further ensuring the adaptivity of prompt structure. The overall framework of \ourapproach{} is shown in Figure \ref{fig:tot}. 

At the beginning, an initial prompt $P_0$ is generated based on few-shot examples~\cite{zhou2022large}\footnote{Our approach can also optimize human-written prompts. In this paper, we use LLM-based prompt initialization to facilitate human effort.}.
Then, we start to iteratively optimize the prompt by using failed cases as feedback. 
%evaluate $p_0$ on $D_{train}$ and sample $k$ failure cases. 
Details of the algorithm can be found in Algorithm \ref{algorithm}. 
Below, we provide a detailed introduction to each module.

%Concretely, for each failed case, the \textit{LLM-Analyzer} conducts error analysis and identify the underlying causes for the current prompt. 

%Then the \textit{LLM-Summarizer} summarizes the causes into different patterns and assigning an importance score to each pattern.
%At last, the \textit{LLM-Revisor} decides whether to expand the branches of the instructions or increase the depth of the instructions, and performs editing based on the analysis results. Finally, the best prompt is selected based on the evaluation metric function $R$.

% \begin{algorithm}
% \caption{Ev}
% \begin{algorithmic}[1]
% \Require Training set; Initial prompt for the target task (i.e., the one we want to refine).
% \State \textbf{// LLM agents}
% \State Analyzer: Conduct error analysis.
% \State Revisor: Expand the tree of instructions.
% \While{$t \leq T$}
%     \State With initial prompt $P_0$ predict each case in $D$ and record failed cases.
%     \State \textbf{//Analyzer}
%     \State Conduct a detailed error analysis based on the failed cases.     \State Identify the key steps where errors occur
%     \State Propose effective solutions.
%     \State \textbf{// Revisor}
%     \State Expand the instruction tree based on expert advice:
%     \State Vertical expansion: Improve steps by including analyzer suggestions.
%     \State Horizontal expansion: Add new branches to handle new patterns.
   
% \EndWhile
% \State \Return The tree prompt.
% \end{algorithmic}
% \end{algorithm}

\begin{algorithm}[t]
\small
\caption{\ourapproach{}}
\begin{algorithmic}[1]
\Require Initial prompt: $P_0$, Training set: $D_{train}$, Validation  set: $D_{val}$, Iteration limit: $T$
\State
\State \textbf{// LLM agents}
\State Analyzer: Conduct error analysis.
\State Summarizer: Condense error reasons into patterns and assign important scores.
\State Revisor: Edit the multi-branched prompt.
\State
\State Initiate $P = P_0$
\While{$t \leq T$}
    \State Evaluate $P$ on $D_{train}$ and sample $K$ failed cases $E = \{(x_i, y_i) | (x_i, y_i) \in D_{train} \land LLM(x_i;p) \neq y_i\}$
    \State \textbf{// Pattern Recognition}
    \State LLM-Analyzer conducts error analysis and identifies the error reasons for each failed case: $R = \{r_1, r_2, \ldots\, r_K\}$
    % \State Select $r_j$, identify error-prone steps $s^*_{r_j} = \arg\max_{s_i \in P_0} \Pr(e \mid s_i, r_j)$
    \State LLM-Summarizer condenses $R$ reasons into $M$ patterns and assign important scores from each pattern: $Patt=\{(patt_1, s_1), \ldots (patt_M, s_M)\}$
    \State \textbf{// Branch Adjustment}
%     \For {\textbf{each} $f_m \in F$} \textbf{parallel for}
%        \State Determine prompt edition mode: adding new branch or enhancing existing.
    \State Selects top $N$ patterns by important scores.
    \State LLM-Revisor optimizes $P$ based on top $N$ $Patt$ and obtained optimized prompts $P_{opt}=\{P1, \ldots P_N\}$.
    \State \textbf{// Branch Pruning}
    \State LLM-Revisor prunes the optimized prompts $P_{opt}$ and obtained $P_{pruned}=\{P1, \ldots P_N\}$
    \State Evaluate $N$ pruned prompts $P_{pruned}=\{P1, \ldots P_N\}$ on $D_{val}$ and select the best prompt $P^*$.
    \State Update $P = P^*$
\EndWhile
\State \Return The best prompt.
\end{algorithmic}
\label{algorithm}
% \vskip -0.15in
\end{algorithm}

\subsection{Pattern Recognition}
Given the current prompt $P$ and $K$ bad examples drawn from the training set $D_{train}$, the goal of this module is to perform error analysis to uncover the root causes of each bad example. To facilitate this, we have created step-by-step instructions for an LLM agent named LLM-Analyzer. The meta prompt for LLM-Analyzer is provided in Appendix \ref{LLM-Analyzer Meta-Prompt}. 
By analyzing the output reasons, we found that the reasons produced by LLM are often similar, even though the failed cases differ. This leads to low optimization efficiency of feedback-based methods.
Furthermore, in AMPO, similar feedback can lead to redundant branches within the prompt, ultimately affecting its performance (as shown in Section~\ref{exp:ablation}).

To address the aforementioned issue, we employ another LLM agent (named as LLM-Summarizer), to summarize the causes of all failed cases into different patterns for each iteration. Incorporating this summarization offers several benefits: (1) It reduces repetitive reasons, significantly enhancing the efficiency of optimization compared to other feedback-based methods. (2) Summarizing reasons into patterns improves generalizability, thereby minimizing redundant branches in the multi-branched prompt. Additionally, we ask the LLM-Summarizer to assign an importance score for each pattern. 
In our paper, we select top $N$ patterns by importance scores.
It enables the selection of important patterns during the branch adjustment process, further reducing the number of explored prompts and thereby improving efficiency.
The meta prompt for the LLM-Summarizer is provided in Appendix \ref{LLM-Summarizer Meta-Prompt}.

% \subsection{Branch Adjustment}
% The goal of the branch adjustment module is to optimize a multi-branched prompt based on the summarized patterns from the LLM-Summarizer. 
% Basically, it needs to first identify which branches of the prompt did not handle the pattern, thereby leading to the failure, and then incorporates the pattern into the multi-branched prompt by making corresponding editions.
% We propose that the key aspect of this process is identifying the \textit{golden mean} of the multi-branched structure, which means creating an adaptive prompt structure based on the distribution of the data.
% It is important to choose an appropriate prompt structure for specific tasks.
% For example, the more complicated the task, the more possible patterns are. In that case, the multi-branched prompt is better because it has better scalability and a larger capacity to handle different patterns. On the contrary, the simpler the task, the better the single flow instructions are, because the single flow of instructions is easier to follow and more robust. 

\subsection{Branch Adjustment}
The goal of the branch adjustment module is to optimize a multi-branched prompt based on the summarized patterns from the LLM-Summarizer. It is critical to clarify whether a pattern should be used to enhance an existing branch or create a new one. 
The optimal choice depends on the specific task. For instance, more complex tasks are likely to exhibit a wider range of patterns, making a multi-branched prompt preferable due to its scalability and capacity to handle various patterns. Conversely, for simpler tasks, single-flow instructions are more effective, as they are easier to follow and more robust. 

\paragraph{Adaptivity between adding the branches to address the new pattern and providing more details to enhance the existing branches}

In this module, we employ an LLM agent, referred to as "LLM-Revisor," to modify the multi-branched prompt. Specifically, we define two types of operations for the LLM-Revisor: (1) adding branches to address new patterns and (2) providing additional details to enhance existing branches. By analyzing the existing branches and new pattern, the LLM-Revisor determines whether to expand the prompt in depth (by adding more details) or in breadth (by adding more branches). 
The advantage of this approach lies in its flexibility, allowing for the prompt structure to be tailored to the task's complexity and pattern distribution. The step-by-step instructions for the LLM-Revisor are provided in Appendix \ref{LLM-Revisor Meta-Prompt}.

\subsection{Branch Pruning}

By thoroughly examining the edition process of the LLM-Revisor, we observed that the multi-branched prompt is susceptible to overfitting. Typically, after several rounds of iteration, the performance of the prompt on the test set begins to decline. Meanwhile, the prompt appears to have more and more branches perceptually.
This happens when the LLM memorizes corner cases in the training data and fails to pick up essential patterns. 
%In machine learning, decision tree pruning is a critical technique used to optimize tree models by reducing overfitting and improving generalization to new data. 
Inspired by pruning techniques in machine learning \cite{esposito1997comparative, kwon2022fast}, we propose two possible solutions: (1) \textit{pre-pruning} prevents the prompt from further growth by early stopping. After each iteration, we check the cross-validation error. If the error does not decrease significantly then we stop. By pruning early, we obtain a more streamlined prompt that is less prone to overfitting the training data.
(2) \textit{post-pruning} does the opposite of pre-pruning and allows the prompt to grow to its full depth.  Particularly, we add a step at the end of the meta prompt for the LLM-Revisor to review the entire set of instructions again and delete any branches to enhance the instructions' generalization ability.

\section{Experimental}

\subsection{Baselines}
% MI(manual instruction), CoT, APO, Prompt Agent, Ours, 
\textbf{Human Instruction}: Human prompts are the instructions crafted by humans based on their understanding and cognition of the original dataset. For each task, we take instructions written by domain experts, sourced from scholarly databases or professional websites, and make corresponding modifications to adapt them to our tasks.
%We employs original prompts from the original dataset or references the work of \citet{guo2023connecting}.

\textbf{Chain-of-Thought (CoT)} \cite{wei2022chain} appends "Let's think step by step." after the question to trigger the model's reasoning process.

\textbf{Chain-of-Thought Instructions (CoT Instructions)} \cite{wei2022chain}: We randomly select 5 cases and generate corresponding thought chains by few-shot learning. After that, we manually optimize the prompts to enhance performance. This prompt is also used as the initial prompt for other feedback-based prompt optimization methods.

\textbf{APO} \cite{pryzant2023automatic}: APO generates natural language-level gradients from incorrect examples, and then utilizes these gradients to reverse-edit the prompt.

\textbf{OPRO} \cite{yang2023large}: OPRO leverages historical prompts, scores, and error examples to guide the LLM in generating higher-scoring prompts. Unlike APO, OPRO does not provide explicit feedback during the optimization process.

\textbf{PromptAgent} \cite{wang2023promptagent}: PromptAgent utilizes the Monte Carlo Tree Search  (MCTS) algorithm to strategically optimize the prompting process.
%generate multiple candidates prompts for the optimial error effedback to maximize futreu rewards. 

% 我们将我们的方法和以下方法进行比较：（a）手工指令：采用源数据集中的prompt或者参考guo等人的工作。（b）CoT：我们为每个任务随机挑选5个用例生成思维链，并进行必要的手工校正。另外，我们也使用了CoT的zero-shot版本，仅在问题之后添加“Let's think step by step.”触发推理过程。
%此外，我们也挑选了2个先进的基于反馈的prompt优化方法。（c）APO：使用错误数据生成描述错误原因的反馈信息，并使用它进行prompt编辑。（d）PromptAgent：同样使用反馈信息进行prompt编辑，与APO不同的是，Promptagent结合蒙特卡洛树搜索细化prompt的优化过程。
% In our research, we compare our method \textcolor{red}{(replace)} with following approaches:

% \textbf{(a) Manual instructions (MI)}: This approach employs original prompts from the source dataset or references the works of \cite{guo2023connecting}. \textbf{(b) Chain of Thought (CoT)} \cite{wei2022chain}: For each task, we randomly select five cases and generate corresponding thought chains. These chains are manually corrected post-generation to ensure quality. Additionally, we explore a zero-shot version of CoT by appending "Let's think step by step." after the question to triggers the model's reasoning process.

% We also compare 2 state-of-the-art feedback-based prompt optimization methods:

% \textbf{(c) APO} \cite{pryzant2023automatic}: This method utilizes feedback detailing the reasons for errors from incorrect responses to edit prompts. \textbf{(d) PromptAgent} \cite{wang2023promptagent}: Similar to APO but enhancing the optimization process through the use of Monte Carlo Tree Search (MCTS) for prompt refinement.

\subsection{Tasks}
To thoroughly evaluate the effectiveness of our method across a wide range of applications, we carefully select 5 tasks from different domains for in-depth experimentation: the well-known text classification task TREC \cite{voorhees2000building}, the widely recognized sentiment classification task SST-5 \cite{socher2013recursive} and the large-scale reading comprehension task RACE \cite{lai-etal-2017-race}. Additionally, we chose two domain-specific tasks from the biomedical field, which explicitly require domain insights when crafting expert-level prompts, namely the medical question-answering tasks MedQA \cite{jin2021disease} and MedMCQA \cite{pal2022medmcqa}.
Detailed dataset information is available in Appendix \ref{Data split}.

\subsection{Implementation Details}
% We choose GPT-3.5-turbo and GPT-4 as evaluators and consistently use GPT-4 as the optimizer. For predictions, we set the base LLM's temperature to 0.0, and for other cases. In the interest of fairness, APO, PromptAgent, and our method all start with the same initial prompt. During the iterative process, we extract a subset of the test set to serve as a validation set, using its results to assess prompt quality. In the final testing phase, we identify and report the best-performing prompt from the top two scoring candidate prompts.

%我们选择GPT-3.5-turbo和GPT-4作为评估器，使用GPT-4作为优化器。对于Analyzer，为了更全面地采集潜在的错误反馈，我们设置温度为1。对于Revisor，我们设置温度为0，以实现更准确的编辑。在APO，PromptAgent的迭代过程中，我们选择测试集中的部分数据作为验证集来衡量当前prompt的质量，最终选择验证集中分数最高的3个prompt进行最终评估并报告最高分数。

In our study, we utilize GPT-3.5-turbo and GPT-4-turbo as the target model, with GPT-4-turbo serving as the optimizer. To comprehensively capture potential error feedback, we set the temperature parameter of the Analyzer to 1. For the Revisor, we set the temperature to 0 to ensure precision in the edits. We sample $K$=5 bad cases and select the top $N=1$ pattern for LLM-Revisor to optimize. Therefore, we keep only one prompt to iterate. Throughout the iterative processes of APO, PromptAgent and \ourapproach{}, we sample 10\% of the training data as the validation set to assess prompt effectiveness. In our experiments, we run each experiment three times and report the average of the results evaluated on the test set.

\begin{table*}[h]
\small
\centering
\resizebox{1\textwidth}{!}{
\begin{tabular}{clccccccc}
\toprule
\multicolumn{1}{l}{\multirow{2}{*}{\textbf{LLMs}}} & \multirow{2}{*}{\textbf{Methods}} & \multicolumn{3}{c}{\textbf{General NLU}}       & \multicolumn{2}{c}{\textbf{Domain Knowledge}} \\
\cmidrule(lr){3-5} \cmidrule(l){6-7} 
       &     & \textbf{SST-5} & \textbf{TREC} & \textbf{RACE} & \textbf{MedQA}  & \textbf{MedMCQA}    \\ 
\cmidrule{1-7}
\multirow{6}{*}{\makecell{GPT-3.5\\-turbo}}     
    & Human   & 51.56  &69.60& 79.25& 61.25        & 45.75   \\
    & CoT     & 50.00& 63.00 & 77.75&50.50     & 47.25\\
    & CoT Instructions  & 49.56& 67.75& 78.25& 68.25        & 48.25  \\
    
    & APO  & 52.00& 69.00& 78.00& 72.25      & 48.00 \\
    & OPRO & 52.44 & 70.50 & 78.75 & 70.50 & 46.75\\
    & PromptAgent & 54.22& 72.50& 80.75& 71.75   & 47.50\\
    
     & \cellcolor{gray!20}OURS     & \cellcolor{gray!20}\textbf{55.78}\up{1.56} & \cellcolor{gray!20}\textbf{76.00}\up{3.50} & \cellcolor{gray!20}\textbf{81.75}\up{1.00} & \cellcolor{gray!20}\textbf{76.50}\up{4.25} & \cellcolor{gray!20}\textbf{48.75}\up{0.50}  \\   
     \midrule
\multirow{6}{*}{GPT-4-turbo }      
    &  Human  & 52.34& 70.50& 89.75 & 64.50  & 65.75  \\
     & CoT      & 53.86& 63.50& 88.50& 63.50        & 69.75\\
    & CoT Instructions & 50.33& 71.25& 91.00& 71.75    & 70.75  \\
    & APO & 55.25& 75.25& 90.75& 83.25   & 71.50  \\
    & OPRO & 56.44 & 79.50 & 90.00 & 76.50 & 66.00 \\
    & PromptAgent & 57.33& 81.50& 91.00& 77.00  & 70.25  \\
     & \cellcolor{gray!20}OURS     & \cellcolor{gray!20}\textbf{59.78}\up{2.45} & \cellcolor{gray!20}\textbf{82.00}\up{0.50} & \cellcolor{gray!20}\textbf{91.25}\up{0.25} & \cellcolor{gray!20}\textbf{89.00}\up{5.75} & \cellcolor{gray!20}\textbf{73.00}\up{1.50}  \\   
          
\bottomrule
\end{tabular}
}
\caption{Performance comparison of GPT-3.5-turbo and GPT-4-turbo across five tasks, highlighting the highest accuracy results in bold. The up arrow indicates the amount by which OURS exceeds the second-highest score.}

\label{tab:main_others}
\end{table*}

\subsection{Main Results}
In Table \ref{tab:main_others}, we present a comparison of prompts generated by \ourapproach{} against Human Instruction, CoT, CoT Instructions, APO, OPRO, and PromptAgent across five tasks in three domains. We observed that our method significantly outperforms accuracy in all tasks, which validates the effectiveness of our approach in optimizing prompts. 
\paragraph{\ourapproach{} significantly surpasses other methods in complex tasks}
Taking the MedQA task as an example, this task includes various conditions and complex situations. 
Therefore, when addressing such issues, it is crucial for LLMs to identify different patterns based on the specific conditions of patients and provide the most suitable treatment plan. 
As shown in Table \ref{tab:main_others}, whether using GPT-3.5-turbo or GPT-4-turbo, the human-constructed prompt performs the worst, possible because the instructions are general and do not align closely with the input information. Then, prompts generated through few-shot examples perform better. Next, when supplemented with failure cases as feedback, APO, OPRO and PromptAgent can further improve the performance. Finally, our method achieved relative improvements of 24.50\%, 25.50\%, 17.25\%, 5.75\% and 12.00\% compared to the other methods (i.e., Human Instruction, CoT, CoT-Instructions,  APO, OPRO, and PromptAgent) using GPT-4-turbo. 
Because our method offers a multi-branched prompt to handle complex situations by categorizing the problems first rather than using a single flow to address all issues. 
The experimental results demonstrate that \ourapproach{} significantly surpasses other methods in complex tasks.
\begin{table}[t]
    \centering
    \small
 \begin{tabular}{lccc}
    \hline
    \textbf{Model} & \textbf{MedQA} & \textbf{TREC} & \textbf{SST-5} \\
    \hline
    \textbf{\ourapproach{}} & \textbf{89.00} & \textbf{82.00} & \textbf{59.78} \\
    \quad - Summarization & 86.75 & 81.50 & 57.00  \\
    \quad \quad \textbf{$\Delta$} & -2.25\% & -0.50\% & -2.78\% \\
    \hline
    \quad - Enhance existing & \multirow{2}{*}{86.25} & \multirow{2}{*}{78.25} & \multirow{2}{*}{55.33} \\
    \quad \quad branches & & & \\
    \quad \quad \textbf{$\Delta$} & -2.75\% & -3.75\% & -4.45\% \\
    \hline
    \quad - Add new branches & 82.25 & 77.75 & 53.78 \\
    \quad \quad \textbf{$\Delta$} & -6.75\% & -4.25\% & -6.00\% \\
    \hline
\end{tabular}

    \caption{The ablation study results of \ourapproach{} without summarization, enhancing existing branches and adding new branches. The exact match score is reported.}
    \label{tab:ablation_study}
     \vskip -0.15in
     
\end{table}

\paragraph{\ourapproach{} also outperforms the other methods in normal tasks}
Take a relatively normal task RACE as an illustration.
As we can see from Table \ref{tab:main_others}, the prompts constructed by human and few-shot examples both perform well. Then, the improvements brought by using feedback-based methods like APO and PromptAgent are minimal, or even detrimental. For example, on the RACE task, the APO-optimized prompt performs 0.25\% lower than the initial prompt (i.e., CoT-Instructions) with GPT-4-turbo as the target model.
Meanwhile, \ourapproach{} consistently surpassed other methods and achieved state-of-the-art performance. It indicates that our method performs well even under normal circumstances.
This means that our method can flexibly create an adaptive multi-branched prompt from the data distribution, thereby adapting to complex or normal tasks.
% 1. 主实验结果

% analysis
% 1. 探索次数 5次  apo
% 2. 聚类的分析。 feedback 的筛选规则：random（聚类之后随机选一个）， greedy（1）， beam search（2，3,5）
% 3. 不聚类，随机选2,3,5个 pk 聚类 beam search（2，3,5）

% Case study： 模仿Promptagent figure5 画个字数少的图

\subsection{Ablation Study} \label{exp:ablation}

To systematically study the effects of the pattern summarization, adaptivity of multi-branched and branch pruning in \ourapproach{}, we conduct thorough ablation experiments across three tasks. The results are shown in the Table \ref{tab:ablation_study}.
\paragraph{Summarization} For each iteration of \ourapproach{}, we incorporate the LLM-Summarizer to summarize the error reasons of all K sampled failed cases into M patterns. By removing the LLM-Summarizer, the error reasons produced by LLM-Analyzer are directly fed into the LLM-Revisor in batch.  According to the results in the Table \ref{tab:ablation_study}, in the MedQA, TREC, and SST-5 tasks, the performance respectively decreased by 2.25\%, 0.50\%, and 2.78\%. It indicates that it is crucial to summarize the error reasons to common patterns before revising.
\paragraph{Adaptive Adjustment} One of the significant innovations of \ourapproach{} is its adaptivity between expanding new branches and enhancing the existing branches of a multi-branched prompt, making it necessary to validate the importance of this feature through ablation experiments. 
Specifically, we modified the meta prompt of LLM-Revisor by removing options to enhance existing branches or adding new branches. 
The experimental results show that the performance significantly decreases by an average of 5.67\% across three tasks without adding new branches. Notably, by removing the addition operation of new branches, \ourapproach{} would actually degrade into APO.
When LLM-Revisor can only add new branches, the performance also decreases by an average of 3.65\%, but still remains higher than enhancing existing branches-only by 3.03\%. 
%This demonstrates that our \ourapproach{}, by enhancing branching capabilities and introducing new sub-branches on the main branch, activates the LLM's capacity for categorical processing, particularly in tasks involving specialized domain knowledge, enabling better handling of complex and varied problems.
Through this study, we have arrived at two conclusions:
(1) both adding new branches and enhancing existing ones are crucial for developing an adaptive multi-branched prompt; (2) adding new branches is more significant than enhancing existing ones when handling complex tasks.
\section{Analysis}

% \subsection{Cluster}

% In carrying out various tasks, we often encounter a large number of bad cases, and consequently, the feedback generated from these cases is also plentiful. It’s important to note that our method of generating feedback differs from the APO's fixed quantity requirements, which could lead to arbitrarily fabricated reasons. Instead, we allow the LLM to autonomously analyze the reasons for errors, resulting in more rational causes. To systematically explore the effect of clustering on our method, we conduct thorough ablation studies with both clustered and non-clustered feedback. For non-clustered scenarios, we additionally experiment with varying numbers of sampled feedback to compare their effectiveness. The strategies for exploring clustering will be more deeply investigated in subsequent chapters.
\begin{figure}[t]
    \centering
    \includegraphics[width = \linewidth, trim=0cm 0cm 0cm 0cm, clip=true]{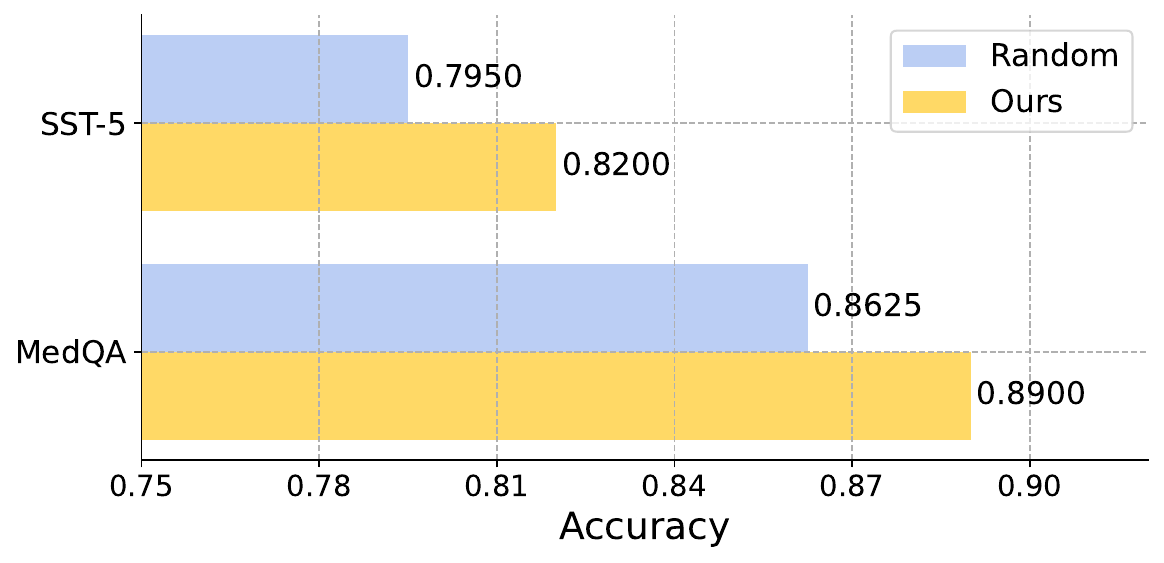}
    \caption{ Comparison between our pattern selection strategy and the random strategy on MedQA and SST-5.}
    \label{fig:Strategy}
     \vskip -0.15in
\end{figure}

\begin{figure}[t]
    \centering
    \includegraphics[width = \linewidth]{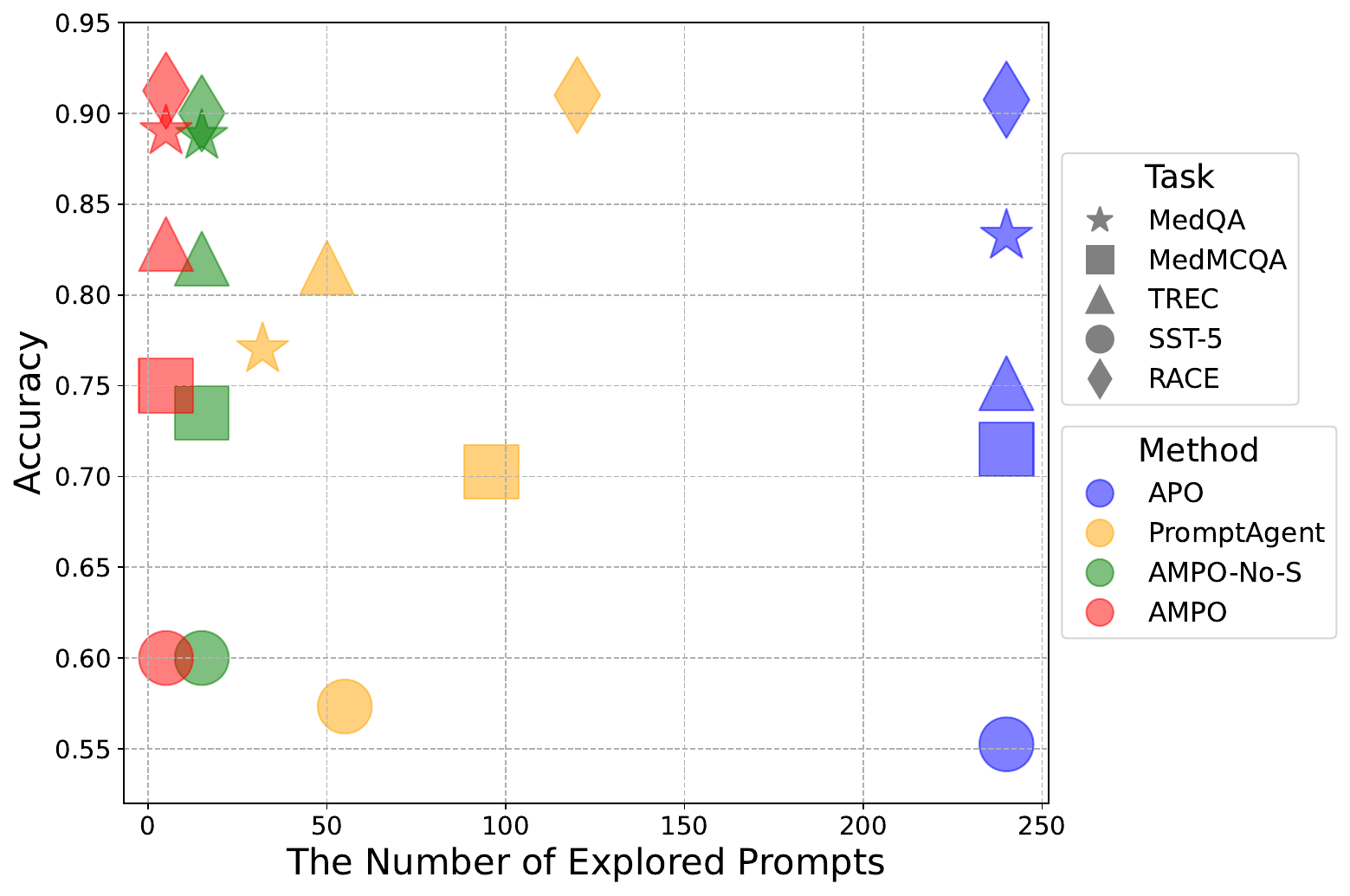}
    \caption{Exploration efficiency analysis. Our method achieved the best results with the fewest exploration prompts. The horizontal axis represents the number of intermediate exploratory prompts, while the vertical axis represents accuracy. Here, \ourapproach{}-No-S refers to \ourapproach{} without the Summarizer.}
    \label{fig:prompt_nums}
     \vskip -0.15in
\end{figure}

\subsection{Pattern Selection Strategy}

In our experimental design, we start by randomly sampling $K$=5 bad cases and analyzing them with the LLM-Analyzer, without specifying a fixed number of causes. After that, all identified reasons are handed over to the LLM-Summarizer, which consolidates them into several main categories. Meanwhile, we ask the LLM-Summarizer to assign an importance score for each main reason. After that, we select the top $N$=1 most critical pattern by important scores. Figure~\ref{fig:Strategy} shows that our search strategy has improved by 2.75\% over random sampling on average, which demonstrates the effectiveness of our pattern selection strategy.
\begin{figure}[t]
    \centering
    \includegraphics[width = \linewidth, trim=0cm 0cm 0cm 0cm, clip=true]{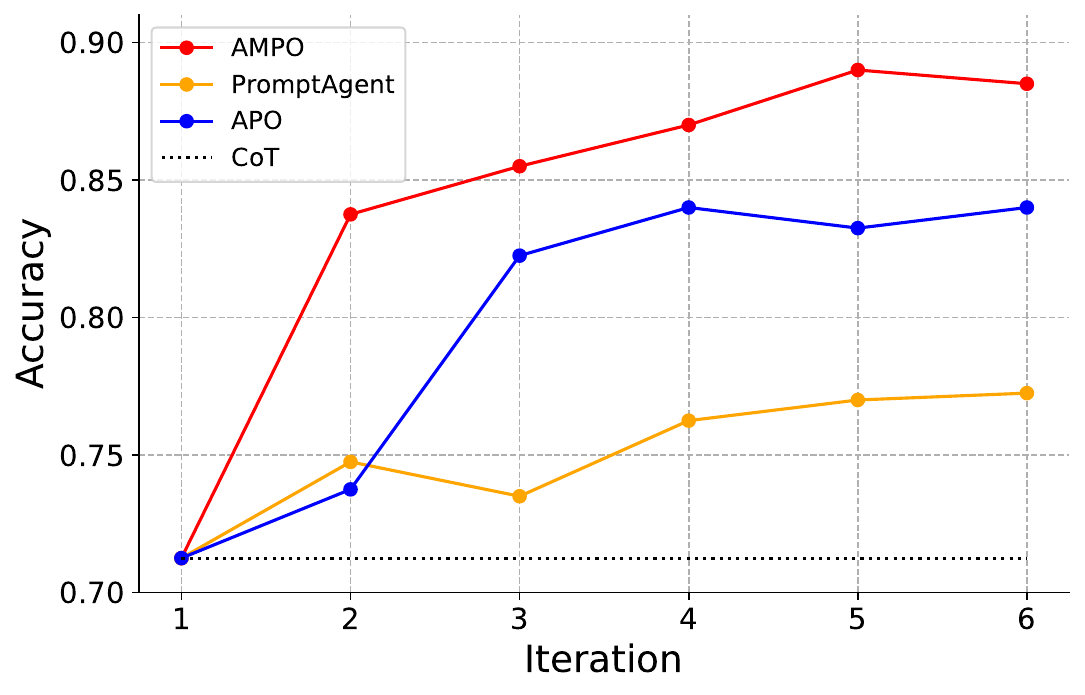}
    \caption{Convergence Analysis}
    \label{fig:convergence}
     \vskip -0.15in
\end{figure}

\begin{figure*}[t]
    \centering
    \includegraphics[width=\textwidth]{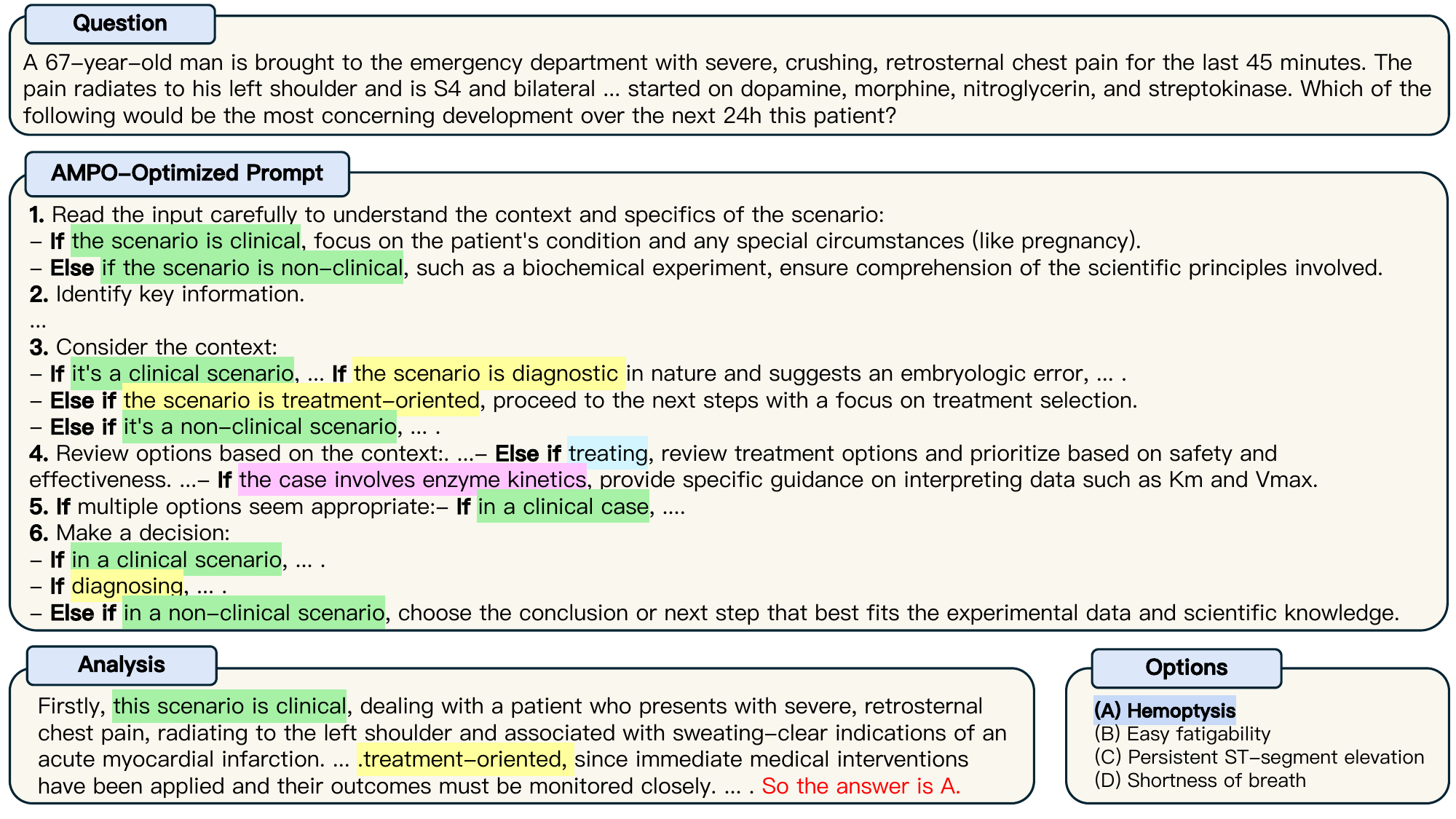}
    \caption{Here is an example from MedQA where the \ourapproach{}-optimized prompt led the LLM to make the correct choice, while other methods all failed. Intuitively, the multi-branched prompt generated by \ourapproach{} has branches at different steps, considering various conditions with if-else statements. 
    Compared to a single flow instruction, it can handle a wider variety of patterns, thereby achieving better performance. We use different colors to highlight various judgment conditions of our prompts. From this example, the multi-branched prompt first guides the LLMs to differentiate problems into \hlgreen{clinical} and \hlgreen{non-clinical}, further judge whether they are \hlyellow{diagnostic} or \hlyellow{treatment-oriented}, while also assessing whether the patient is suitable for treatment. For cases \hlcustpurple{involving enzyme kinetics},  it will guide the LLMs to provide insights specific to that field.}
    \label{fig:case_study}
     \vskip -0.15in
\end{figure*}

\subsection{Exploration Efficiency Analysis}
Exploration efficiency is critical to reduce the computation cost.
We thus compare the number of explored prompts between our method
and three strong baselines. In Figure \ref{fig:prompt_nums}, ours achieves the best results with the fewest exploration prompts across various tasks.
Take the MedQA task as an example, we performed 5 iterations for 4 different methods, calculating the number of prompts they generated. For each iteration, APO generates 3 prompts from each of the 4 prompts of the previous iteration, and selects the 4 prompts with the best performance on the validation set for the next iteration, resulting in 240 exploration prompts. PromptAgent, using an MCTS Search strategy\cite{winands2008monte} with a depth of 3 and a breadth of 3, produces a total of 52 prompts after 5 iterations. Compared to PromptAgent and APO, \ourapproach{} uses 6.4 times and 48 times fewer explored prompts respectively, yet achieved performance improvements of 12\% and 5.75\%. 
The high efficiency of our method is mainly due to three reasons: (1) We employ a greedy search strategy, meaning that in each iteration, only one best prompt is retained. (2) Our LLM-Summarizer condenses all the error reasons from failure cases into several patterns. (3) Additionally, the LLM-Summarizer assigns an important score for each pattern, allowing the LLM-Revisor to filter the patterns, which further reduces the number of exploration prompts.

\subsection{Convergence analysis}
To further investigate the learning process of \ourapproach{}, we monitored and visualized the performance changes of prompts over each round of the process. Specifically, we recorded and plotted the performance trajectories of all baselines across six rounds in the MedQA task, illustrating the evolution of the prompt optimization methods' performance in Figure \ref{fig:convergence}. We observed that all three optimization methods showed an overall upward trend, but \ourapproach{}'s increase is notably greater, jumping directly from 71.25\% to 83.75\% only one iteration.  
The multi-branched prompt has better scalability and a larger capacity to handle different patterns. 
Unlike other methods, our approach can expand new branches besides modifying existing ones, allowing it to handle more situations and reducing the difficulty of modifying original prompts.

\subsection{Case Study}
To illustrate how \ourapproach{} utilizes multi-branched prompts to solve issues, we conducted a qualitative analysis. By using an example from MedQA, we demonstrated that our approach can accurately categorize complex scenarios from intricate data. Then it can design detailed solutions for each case, ultimately leading to the correct answer. From the analysis result of this example, we can see that the LLM first judges the patient's condition as clinical and finds that there are clear indications of an acute myocardial infarction. Next, it judges the situation as treatment-oriented and finally arrives at the correct answer.
% \section{Discussion}

% In this section, we discussed principles to choose between COT prompt and tree structure prompt, and some issues found in the process of building tree structure prompts, such as being prone to over-fitting and having redundant branches. For these issues, we analyze the reasons and discuss possible solutions in depth.

\section{Conclusion}
In this paper, we proposed Automatic Multi-Branched Prompt Optimization (AMPO), which explicitly converts a prompt into a multi-branched format and then iteratively refines it using failure cases as feedback. Specifically, we employ three LLM agents working in tandem and propose guiding principles to balancing the adaptability of the multi-branched structure. Experimental results demonstrate that \ourapproach{} outperforms existing state-of-the-art feedback-based optimization approaches while significantly improving optimization efficiency.

\section{Limitations}
Multi-branched prompt optimization requires LLMs to possess strong logical reasoning abilities. To reduce complexity, we have adopted the divide and conquer approach, designing three roles: LLM-Analyzer, LLM-Summarizer, and LLM-Revisor. Additionally, we have designed step-by-step meta instructions to guide how to generate an adaptive prompt structure to accommodate tasks of varying difficulties. Despite this, our method still depends on the capabilities of the LLMs themselves. Due to current limitations in their abilities, there are times when the models may not strictly follow the meta instructions, leading to suboptimal results. However, by utilizing better LLMs in the future, we can further enhance the effectiveness of our method.

\bibliography{custom}

\appendix
\section{Appendix} 
\subsection{Data split} \label{Data split}
In our experimental setup, tasks are organized into two primary categories: General NLU (Natural Language Understanding) and Domain Knowledge.

Within the General NLU category, we have three tasks: SST-5, TREC, and RACE. Each task is allocated 100 training samples and 50 evaluation samples. For the test sets, SST-5 comprises 450 samples, while TREC and RACE each have 400 samples.

For the Domain Knowledge category, there are two tasks: MEDQA and MEDMCQA. Like the General NLU tasks, each of these tasks is assigned 100 training samples. However, the table indicates that there are 50 evaluation 
\begin{table}[h]
\footnotesize
    \small
\centering
\renewcommand{\arraystretch}{1}
\setlength\tabcolsep{4pt}
\begin{tabular}{lllll}
\toprule
\bf{Type}                                      & \bf{Task}                            & \bf{Train }                                 & \bf{Eval}                                                        & \bf{Test}                     \\
\midrule
\multirow{9}{*}{General NLU}      & \multirow{3}{*}{SST-5}        & \multirow{3}{*}{100}       & \multirow{3}{*}{50}                 &                     \\
                                          &                                   &                                           &                                                                  & 450                 \\
                                          &                                   &                                           &                                                                  &                        \\
                                           \cmidrule{2-5}
                                          & \multirow{3}{*}{TREC}           & \multirow{3}{*}{100}  & \multirow{3}{*}{50}        &                    \\
                                          &                                   &                                           &                                                                  & 400                   \\
                                          &                                   &                                           &                                                                  &                       \\
                                          \cmidrule{2-5}
                                          & \multirow{3}{*}{RACE}        & \multirow{3}{*}{100} & \multirow{3}{*}{50}        &                   \\
                                          &                                   &                                           &                                                                  & 400                      \\
                                          &                                   &                                           &                                                                  &                        \\
                                          \midrule
\multirow{6}{*}{Domain Knowledge} & \multirow{3}{*}{MEDQA} & \multirow{3}{*}{100}         &                                                                  &                         \\
                                          &                                   &                                           & 50             & 400                  \\
                                          &                                   &                                           &                                                           &                         \\
                                          \cmidrule{2-5}
                                          & \multirow{3}{*}{MEDMCQA} & \multirow{3}{*}{100}         &                                                                  &                         \\
                                          &                                   &                                           & 50             & 400                  \\
                                          &                                   &                                           &                                                           &                         \\
                                          \bottomrule
\end{tabular}
\caption{\label{tab:def}Experimental Data Distribution}
\end{table}
samples for these tasks samples for these tasks and 400 test samples. It should be noted that there are some formatting inconsistencies in the Eval and Test columns for the MEDQA and MEDMCQA tasks which need to be addressed for clarity. For detailed information, please refer to Table \ref{tab:def}.

\subsection{Different Pattern Results} \label{Pattern}
We explored the impact of selecting the top $N$ patterns on the results. As evidenced by the table, accuracy peaks when $N$ = 5, while it is lowest at $N$ = 4, exhibiting an overall oscillatory trend within a reasonable range. Consequently, to enhance our efficiency, we opted for $N$ = 1.
\begin{figure}[t]
    \centering
    \includegraphics[width = \linewidth, trim=0cm 0cm 0cm 0cm, clip=true]{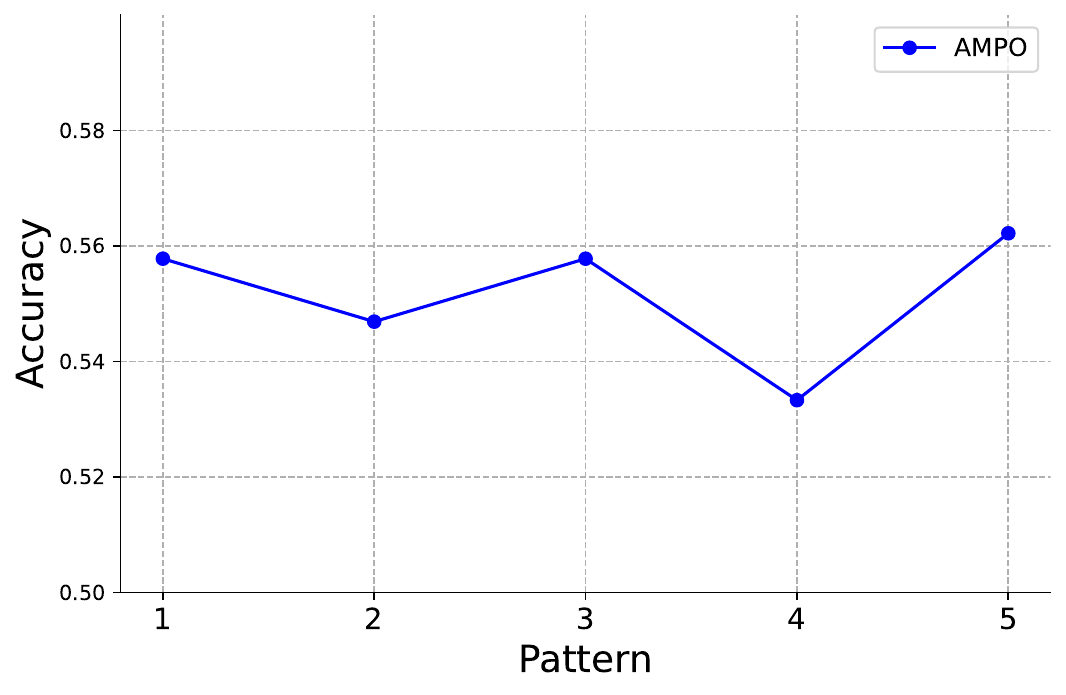}
    \caption{The performance of choosing different numbers of top patterns on SST-5 task. We use GPT-3.5-turbo as the target model.}
    \label{fig:Pattern}

\end{figure}

\subsection{LLM-Analyzer Meta-Prompt} \label{LLM-Analyzer Meta-Prompt}
\begin{table*}[h!]
    \centering
    \begin{tabular}{@{} >{\raggedright\arraybackslash}p{\textwidth} @{}}
    \toprule
        \verb|---|ProblemStart\verb|---| \\
        I have some instructions for a specific problem: \\
        \verb|---|InstructionsStart\verb|---| \\
        \{\{initial\_prompt\}\} \\
        \verb|---|InstructionsEnd\verb|---| \\
        \\
        But it gets the following cases wrong: \\
        \verb|---|BadCasesStart\verb|---| \\
        \{\{bad\_examples\}\} \\
        \verb|---|BadCasesEnd\verb|---| \\
        \\
        Your task is to identify the underlying causes for my [\# Instructions] as an analyzer. Please follow these steps: \\
        (1) Identify what perspectives there are to consider for my problem. Please think as comprehensively as possible, considering all aspects. \\
        (2) Based on these potential perspectives you identified, analyze the pattern of the failed cases. \\
        (3) Carefully review each step of my [\# Instructions] and identify which step neglects the key information in the pattern, resulting in failure. \\
        (4) Write your reasons and wrap each reason with \textless START\textgreater and \textless END\textgreater. \\
    \bottomrule
    \end{tabular}
    \caption{LLM-Analyzer}
\end{table*}
\subsection{LLM-Summarizer Meta-Prompt}
\label{LLM-Summarizer Meta-Prompt}
\begin{table*}[h!]
    \centering
    \begin{tabular}{@{} >{\raggedright\arraybackslash}p{\textwidth} @{}}
    \toprule
        \verb|---|ProblemStart\verb|---| \\
        I have some instructions for a specific problem: \\
        \verb|---|InstructionsStart\verb|---| \\
        \{\{initial\_prompt\}\} \\
        \verb|---|InstructionsEnd\verb|---| \\
        \\
        Here are some reasons why my current instructions cannot solve some problem : \\
        \verb|---|Reasons\verb|---| \\
        \{\{Reasons\}\} \\
        \verb|---|Reasons\verb|---| \\
        \\
        Your task is to summarize the many reasons provided above into a few major categories and assign an important score for each category. Be careful to eliminate repetitive and similar reasons. Each summarized pattern should be wrapped with \textless START\textgreater and \textless END\textgreater. \\
    \bottomrule
    \end{tabular}
    \caption{LLM-Summarizer}
\end{table*}
\subsection{LLM-Revisor Meta-Prompt}
\label{LLM-Revisor Meta-Prompt}

\begin{table*}[h!]
    \centering
    \begin{tabular}{@{} >{\raggedright\arraybackslash}p{\textwidth} @{}}
    \toprule
        \verb|---|ProblemStart\verb|---| \\
        You have some instructions for a specific task:  \\
        \verb|---|InstructionsStart\verb|---| \\
        \{\{initial\_prompt\}\} \\
        \verb|---|InstructionsEnd\verb|---| \\
        \\
       However, due to the complexity of real-world situations, a single flow of instructions (i.e., sequential instructions) cannot apply to all cases. Therefore, you should transform the instructions into a conditional approach, which means adopting different instructions for different patterns.  \\
        \\
        Notably, the key aspect of this process is to create an adaptive prompt structure, thereby accommodating tasks of varying difficulties.
       To achieve this, you should find the \textit{golden mean} between adding the branches to address the new pattern and providing more details to enhance the existing branches based on the difficulty of your task and the distribution of recognized patterns. \\
        \\
        An expert has pointed some patterns that you don't considered before for your instructions: \\
        \verb|---|ExpertAnalysisStart\verb|---| \\
        \{\{patterns\}\} \\
        \verb|---|ExpertAnalysisEnd\verb|---| \\
        \\
        Please optimize your [\# Instructions] based on expert analysis step-by-step:\\
        (1) Carefully review each step of your instructions. \\
        \\
        (2) Identify the steps that went wrong due to a lack of key information mentioned in expert analysis. \\
        \\
        (3) For each suboptimal step, you have the following options:\\
        - 3.1 Consider improving the step to include the key information.\\  
        - 3.2 Otherwise, you can also consider adding **sub-steps** using an **if** or **if-else** structure to handle the **new** patterns. Ensure that each substep is specific and avoids vague instructions. \\
        Note that if a step needs to consider multiple situations, break it down into substeps to make it easier to follow. \\
        \\
        (4) Include Tips or Cautions: If merely optimizing existing steps with branches like if-else does not sufficiently to address all aspects, add new tips or cautions to the current instructions to handle different patterns.  \\
        \\
        (5) Maintain the other main steps unchanged from the initial prompt, in order to not lose information. \\
        \\
        (6) At last, review the whole steps and prune the branches to avoid the instructions overfitting. \\
        \\
        Please only output the optimized prompt without anything else.\\
    \bottomrule
    \end{tabular}
    \caption{LLM-Revisor}
\end{table*}

\end{document}